\documentclass{article}
\usepackage{spconf,amsmath,graphicx}
\usepackage{multirow}
\usepackage[table,xcdraw]{xcolor}
\usepackage[normalem]{ulem}
\useunder{\uline}{\ul}{}
\title{FC3DNET: A FULLY CONNECTED ENCODER-DECODER FOR EFFICIENT DEMOIRÉING}

\name{Zhibo Du \qquad Long Peng \qquad Yang Wang$^{\ast}$ \qquad Yang Cao \qquad Zheng-Jun Zha\thanks{*Corresponding author. Email: ywang120@ustc.edu.cn}\thanks{First author email: duzb@mail.ustc.edu.cn}\thanks{This work was supported by the Natural Science Foundation of China No. 62206262.}}
  
\address{University of Science and Technology of China\\
Department of Automation\\
Hefei, China}

\begin{document}
%\ninept
%
\maketitle
\begin{abstract}

Moiré patterns are commonly seen when taking photos of screens. Camera devices usually have limited hardware performance but take high-resolution photos. However, users are sensitive to the photo processing time, which presents a hardly considered challenge of efficiency for demoiréing methods.
To balance the network speed and quality of results, we propose a \textbf{F}ully \textbf{C}onnected en\textbf{C}oder-de\textbf{C}oder based \textbf{D}emoiréing \textbf{Net}work (FC3DNet). FC3DNet utilizes features with multiple scales in each stage of the decoder for comprehensive information, which contains long-range patterns as well as various local moiré styles that both are crucial aspects in demoiréing.
Besides, to make full use of multiple features, we design a Multi-Feature Multi-Attention Fusion (MFMAF) module to weigh the importance of each feature and compress them for efficiency. 
These designs enable our network to achieve performance comparable to state-of-the-art (SOTA) methods in real-world datasets while utilizing only a fraction of parameters, FLOPs, and runtime.

\end{abstract}
% 100 to 150 words
% limited to 6 pages one optional 7th page containing only references/bibliography
\section{Introduction}
\label{sec:intro}
\begin{figure}[tb]
\centering
\begin{minipage}[b]{0.24\linewidth}
\centering
\centerline{\includegraphics[width=4cm]{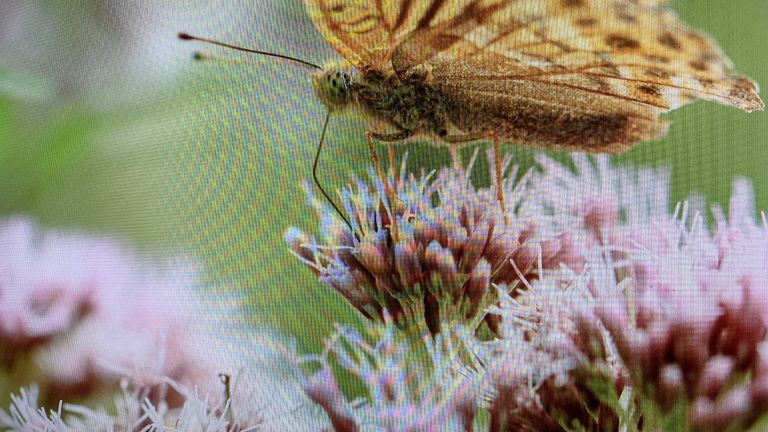}}
 \centerline{(a) Moiré image}\medskip
\end{minipage}
% \hfill
\hspace{2cm}
\begin{minipage}[b]{0.24\linewidth}
\centering
\centerline{\includegraphics[width=4cm]{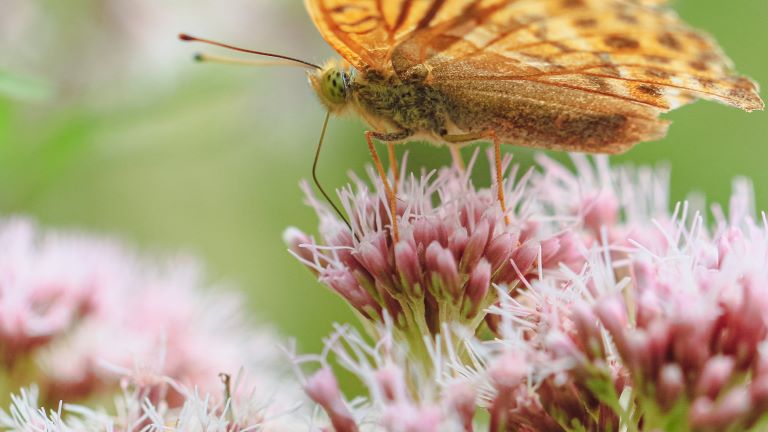}}
% \vspace{2.0cm}
 \centerline{(b) Clean image}\medskip
\end{minipage}

\begin{minipage}[b]{0.16\linewidth}
\centerline{\includegraphics[width=2.6cm]{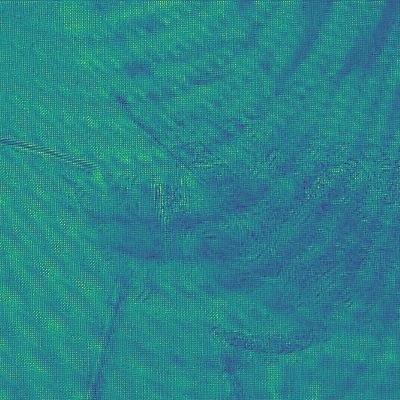}}
 \centerline{(c) Feature $X_{1p}$}\medskip
\end{minipage}
% \hfill
\hspace{1.2cm}
\begin{minipage}[b]{0.16\linewidth}
\centerline{\includegraphics[width=2.6cm]{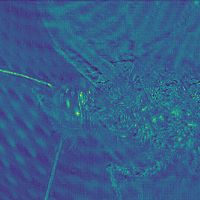}}
\centerline{(d) Feature $X_{2p}$}\medskip
\end{minipage}
% \hfill
\hspace{1.2cm}
\begin{minipage}[b]{0.16\linewidth}
\centerline{\includegraphics[width=2.6cm]{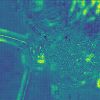}}
 \centerline{(e) Feature $X_{3p}$}\medskip
\end{minipage}

\caption{Example of a moiré image and its ground-truth. The colors and shapes of moiré patterns vary in different areas of the image. $X_{1p}$, $X_{2p}$ and $X_{3p}$ are patches of  $X_1$, $X_2$ and $X_3$, respectively, which are defined in Fig. \ref{fig:nets} and extracted from (a). They represent different characteristics.}
\label{fig:ex}
\end{figure}

Moiré patterns occur due to the frequency aliasing between the pixel grids of the camera sensor and the captured screen. These patterns typically cover the entire image and exhibit varying colors and shapes in different photos and even within different areas of the same image, as shown in Fig. \ref{fig:ex}. Even slight changes in the relative position between the camera and screen can result in dramatic variations of moiré patterns. Initially, some people attempted to remove moiré patterns using non-deep learning methods, but the results were often unsatisfactory. The development of deep learning and neural networks has provided new approaches for image demoiréing and image restoration \cite{wang2023decoupling,wang2023brightness,peng2024lightweight,peng2024towards,peng2024efficient,wang2017deep,li2023ntire,wang2020deep,wang2019progressive,peng2021ensemble,peng2020cumulative,zou2022dreaming}.

Some methods have been proposed based on the detailed prior knowledge of moiré patterns and incorporate complex specific modules.
MopNet \cite{mop} comprises a channel-wise target edge predictor to estimate the edge map of a moiré-free image and an attribute-aware classifier to differentiate between different types of moiré patterns such as frequency, color, and shape. CDFNet \cite{cfd} utilizes similar modules.
In FHDe$^2$Net \cite{fhd}, GDN, LRN, FDN, and FRN are designed and connected complicatedly by the transformation of color space and Discrete Cosine Transform. These networks exhibit slower processing speeds, as shown in Table \ref{table:1}. This is because they decompose the demoiréing task into fine-grained sub-problems and employ specific modules to address them. However, it may be not optimal, leading to learning less important features.

Besides, a commonly employed technique is the multi-scale architecture.
DMCNN \cite{dm} and MDDM \cite{md} utilize multi-scale branches to demoiré at different resolutions, and all outputs are simply added up to obtain the final output.
PMTNet \cite{pmt}, SDN \cite{sdn} and MCFNet \cite{mcf} remove moiré patterns from low scale to high scale. PMTNet incorporates streams in series. MCFNet introduces MGRB in each scale and MGRB also extracts features from low to high.
UNet-like \cite{u} structure has been widely applied for overall architecture or feature extraction.
MBCNN \cite{mb} and ESDNet \cite{esd} are both UNet-like structures. MBCNN employs different blocks at all levels of the structure. ESDNet incorporates pyramid extraction in each level, which extracts features at multiple scales and applies channel attention to fuse them.
MopNet \cite{mop} utilizes a UNet for feature aggregation, where the outputs of each level are concatenated and weighted by SENet \cite{se}. CDFNet \cite{cfd} employs similar modules. FHDe$^2$Net \cite{fhd} incorporates a UNet-like GDN and LRN to remove moiré patterns across all scales. However, most networks only focus on extracting multi-scale features without adequately addressing the fusion of these features appropriately. They fuse features by simplistic addition \cite{dm,md}, SENet \cite{mop,esd} or a convolution layer \cite{mb,pmt}, which can be not enough to integrate image content and moiré information from multiple scales and depths. This can lead to under-utilization, requiring larger networks and more parameters to compensate.

While significant progress has been made in screenshot demoiréing, the aforementioned methods either fail to achieve satisfactory results \cite{dm,md,wd,mop,mb,fhd,pmt,cfd}, or come with a heavy computational cost \cite{mop,fhd,mcf}.
Considering that the majority of people capture photos by smartphones with the hope of a fast processing speed and high-quality images, there is still a pursuit for more lightweight and efficient networks. 
After investigating existing methods, we avoid excessively partitioning the demoiréing problem into sub-problems. Because of the characteristics of moiré patterns mentioned before, we employ the multi-scale strategy to capture long-range patterns as well as local representations, which has been proven to be helpful by many previous works. In contrast to those methods, we pay more attention to the interaction and fusion of multi-scale features.
Overall, we utilize a multi-scale encoder-decoder architecture and design a Feature Full Connection Module (FFCM) to connect the encoder and decoder. When the network needs multi-scale features, they are first fused through a specifically designed feature attention module.

UNet employs skip connections between corresponding levels of the encoder and decoder to allow each level of the decoder to utilize features extracted from its corresponding level of the encoder. The shallower level captures more details and local moiré patterns, while the deeper level captures more abstract information and global moiré patterns due to larger receptive fields, as shown in Fig. \ref{fig:ex}. However, the representations of the same spatial location from different features encode different information about that location of the image. They are not the same and can represent the features of the image more completely only by combining. 
Unlike the UNet-like structure, which is used by many previous works, our encoder consists of three levels in series that extract features from high scale to low scale, and the outputs of every level are concatenated together through FFCM. FFCM generates outputs for each level of decoder and allows them to have more comprehensive multi-scale information from different levels of encoder rather than separating them. Specifically, after removing the connections between features with different scales, FFCM degenerates into skip connections used in UNet.

To make full use of features from different levels and avoid the aforementioned under-utilization problem, it’s necessary to evaluate their importance for the decoder and compress features for efficiency.
We design a Multi-Feature Multi-Attention Fusion (MFMAF) module, which accepts an arbitrary number of features as input and outputs a fusion of them. Inspired by developing methods of attention \cite{se,cbam}, we adopt a lightweight approach to learn the importance of features using channel attention and spatial attention. Different from other works, we have made improvements to the computation details to make them more efficient and suitable for the fusion of multiple features. 

The contributions of our work can be summarized as follows:

\begin{itemize}
  \item We propose an efficient network for demoiréing, which introduces a novel fully connected encoder-decoder that fuse features from every level of the encoder for the utilization by each level of the decoder, which better takes advantage of information from multiple scales compared to previous multi-scale methods. 
  \item Due to recognition of the importance of multi-scale feature fusion, we design a Multi-Feature Multi-Attention Fusion module, which is more suitable for the fusion of multiple features, and it outperforms classic attention-based fusion methods. 
  \item The proposed method achieves comparable performance compared to the state-of-the-art (SOTA) methods while having only a fraction of parameters, FLOPs, and runtime. This significantly strengthens the practicality of deep learning-based demoiréing methods on mobile devices.
\end{itemize}

\section{PROPOSED METHOD}
\label{sec:method}

\begin{figure*}[htb]

\begin{minipage}[b]{1.0\linewidth}
 \centering
 \centerline{\includegraphics[width=17cm]{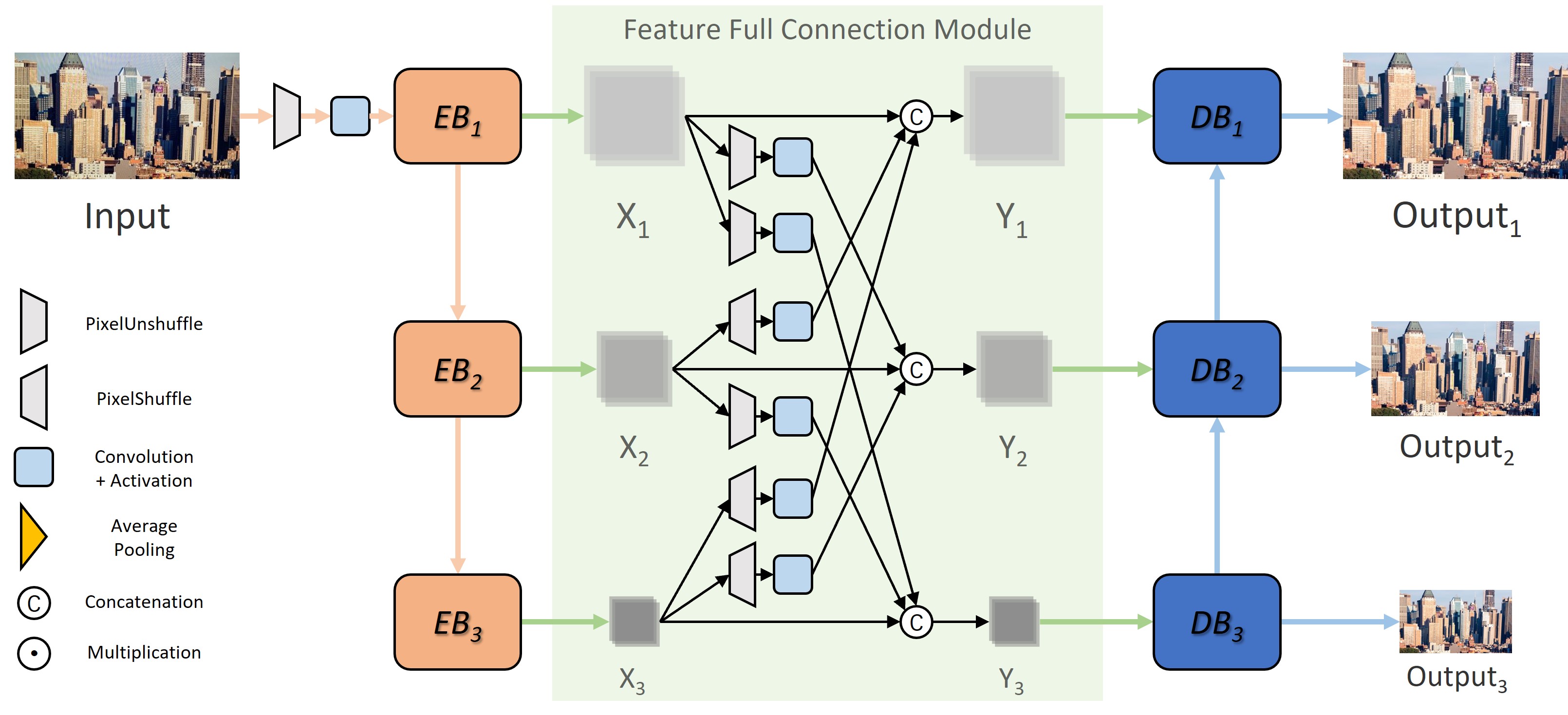}}
% \vspace{2.0cm}
 \centerline{(a) FC3DNet}\medskip
\end{minipage}

\begin{minipage}[b]{0.46\linewidth}
 \centering
 \centerline{\includegraphics[width=7cm]{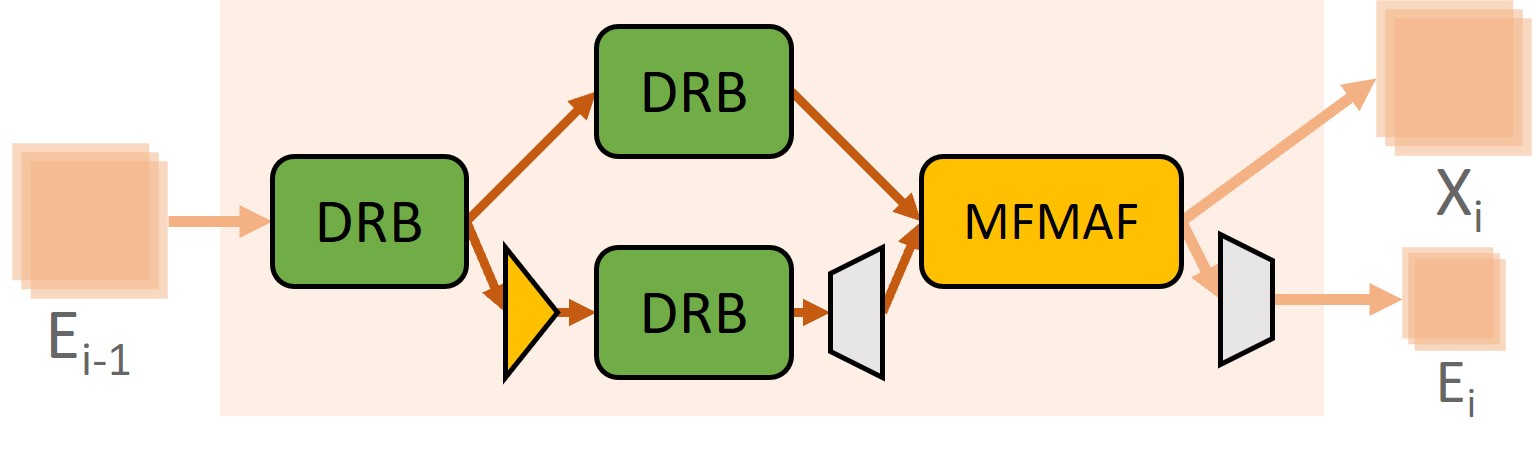}}
% \vspace{2.0cm}
 \centerline{(b) Encoder Block $EB_i$}\medskip
\end{minipage}
\hfill
\begin{minipage}[b]{0.53\linewidth}
 \centering
 \centerline{\includegraphics[width=9.5cm]{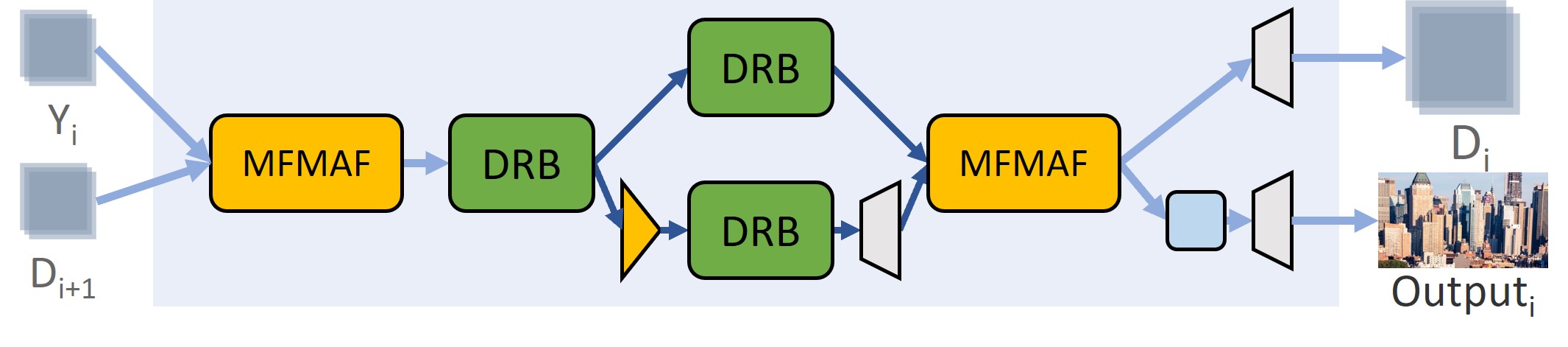}}
% \vspace{2.0cm}
 \centerline{(c) Decoder Block $DB_i$}\medskip
\end{minipage}

\caption{Architecture of the proposed FC3DNet and structure of Encoder Block and Decoder Block.}
\label{fig:nets}
\end{figure*}

\subsection{Overall Network Architecture}
 
The overall architecture, as shown in Fig. \ref{fig:nets}(a), is a fully connected multi-scale encoder-decoder. The input goes through a PixelUnshuffle operation, transforming its size from (3, H, W) to (12, H/2, W/2), where H and W denote the height and width of the input image, respectively. Then a convolution layer extracts its shallow features. The features are sent to the encoder, which consists of three encoder blocks cascaded together, processing features from high scale to low scale. Each block produces an output feature, besides, $EB_1$ and $EB_2$ also generate a feature that will be passed to the next lower-level block. The outputs of three blocks are fed into a Feature Full Connection Module (FFCM), which connects them and generates appropriate features for each level of the decoder.

Similar to the encoder, the decoder comprises three decoder blocks in series, processing features from low scale to high scale. We simply follow \cite{mb,esd} and set three levels. Each block produces an output image with the corresponding size, besides, $DB_3$ and $DB_2$ also generate a feature that will be passed to the next higher-level block. Each block receives an output of FFCM with the corresponding size. The output of $DB_1$ is the moiré-free image that we desire.

To connect all the encoder blocks and decoder blocks, we propose a Feature Full Connection Module, which receives features with multiple sizes and generates features with multiple sizes as well. PixelUnshuffle and PixelShuffle \cite{ps} are applied to resize features from one size to another to enable them to be concatenated together. For different decoder blocks, their requirements for the outputs of different encoder blocks vary. They primarily rely on features at the same level, which processes images in the same scale, while utilizing features from other levels as auxiliary information. Thus convolution layer is applied to adjust the channel number of features, which represents the quantity and importance of information. The operations are formulated as
\begin{equation}
Y_1 = Cat(X_1,CA(PS(X_2,2)),CA(PS(X_3,4))),
\label{eq:c1}
\end{equation}
\begin{equation}
Y_2 = Cat(CA(PU(X_1,2)),X_2,CA(PS(X_3,2))),
\label{eq:c2}
\end{equation}
\begin{equation}
Y_3 = Cat(CA(PU(X_1,4)),CA(PU(X_2,2)),X_3),
\label{eq:c3}
\end{equation}
where $PU$ and $PS$ denote PixelUnshuffle and PixelShuffle, respectively, and there second parameter indicates the downscale or upscale factor. $CA$ denotes a convolution and activation layer. $Cat$ denotes the concatenation operation.
\subsection{Encoder Block}
The Encoder block consists of a feature extraction part and a feature fusion part. After the input $E_{i-1}$ goes through a DRB (Dilated Residual Block), which has two vanilla convolution layers, two dilated convolution layers, and a residual connection between the input and output, it branches into two paths. One path contains a DRB to further extract features at the current scale. In another path, the feature is resized by a factor of 1/3 using average pooling before being passed through a DRB to extract features at a larger receptive field. The output is then adjusted back to the original size using PixelShuffle. The features extracted from two paths are sent to MFMAF for fusion, allowing for comprehensive utilization of information captured at different receptive fields. The output of MFMAF is fed into FFCM and down-sampled with a factor of 1/2 for $EB_{i+1}$ if it exists.

\subsection{Decoder Block}
The decoder block has multiple inputs that need to be fused before further processing, and $D_{i+1}$ is missing when $i$ is equal to 3. We utilize MFMAF to fuse features which isn’t done by FFCM, because it plays a role of connection. After fusing, the operations within the decoder block are the same as those in the encoder block except that the outputs differ slightly. While encoder blocks only output features, decoder blocks produce images as well. The output of MFMAF is restored to a moiré-free image and up-sampled with a factor of 2 for $DB_{i-1}$ if it exists.

\subsection{Multi-Feature Multi-Attention Fusion Module (MFMAF)}

\begin{figure}[t]
\centering
\begin{minipage}[b]{0.48\linewidth}
 \centerline{\includegraphics[width=8.5 cm]{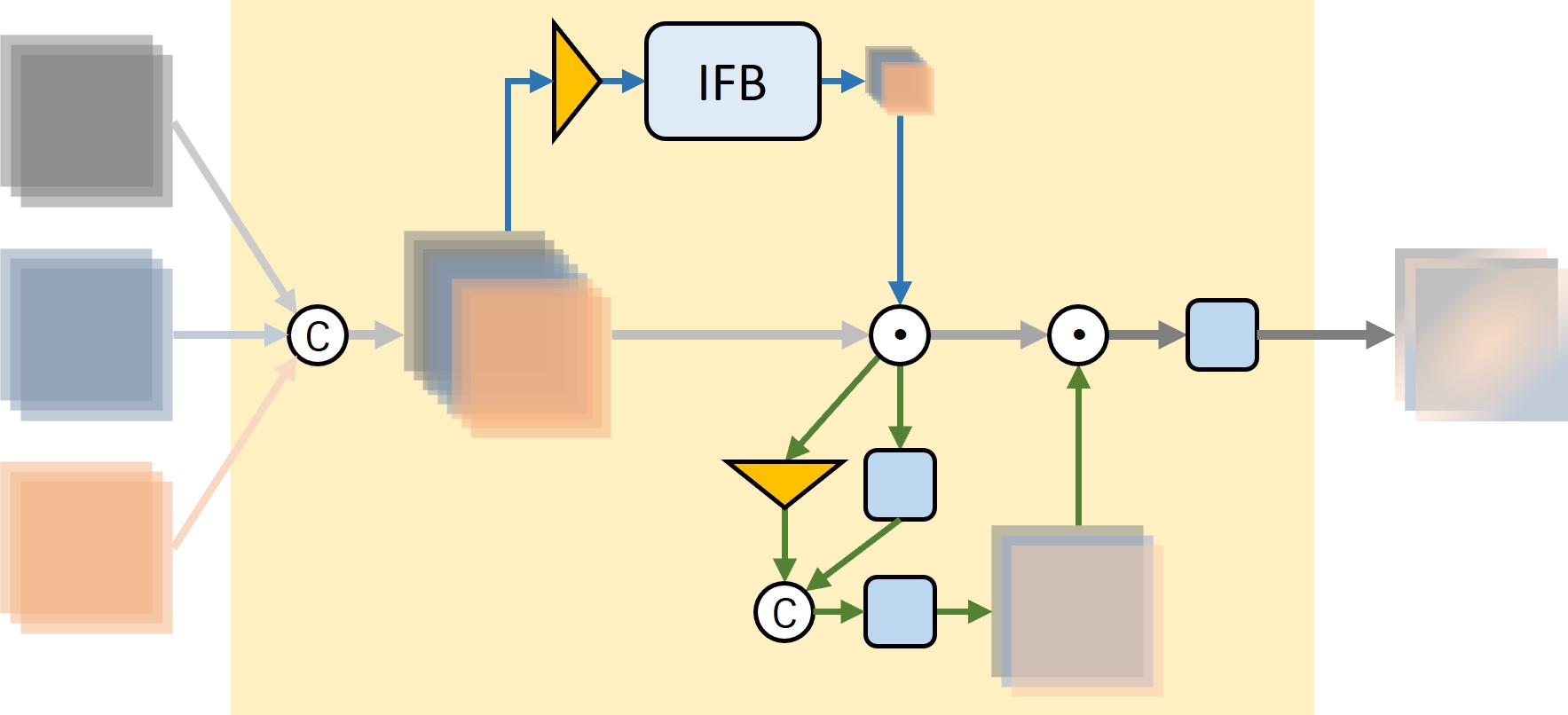}}
\end{minipage}
\caption{Structure of MFMAF.}
\label{fig:fam}
\end{figure}

\begin{table*}[t]
\setlength{\tabcolsep}{3pt}
\centering
\caption{Quantitative comparison between FC3DNet and classic methods on three datasets. FLOPs and runtime are tested on the FHDMi dataset using a RTX 3080 GPU. Results with * are obtaind from a smaller MCFNet. The best result is shown in {\color[HTML]{FF0000}\textbf{bold}}, and the second-best is {\color[HTML]{5B9BD5}{\ul underlined}}.}
\label{table:1}
\scalebox{0.66}{
\begin{tabular}{c|c|ccccccccccc|c}
\hline
Dataset & Metric & Input & DMCNN \cite{dm} & MDDM \cite{mb} & WDNet \cite{wd} & MopNet \cite{mop} & MBCNN \cite{mb} & FHDe2Net \cite{fhd} & PMTNet \cite{pmt} & CFDNet \cite{cfd} & ESDNet \cite{esd} & MCFNet \cite{mcf} & FC3DNet \\ \hline
 & PSNR$\uparrow$ & 17.117 & 19.914 & 20.088 & 20.364 & 19.489 & 21.414 & 20.338 & - & - & 22.119 & {\color[HTML]{5B9BD5} {\ul 22.484}} & {\color[HTML]{FF0000} \textbf{22.487}} \\
 & SSIM$\uparrow$ & 0.5089 & 0.7575 & 0.7441 & 0.6497 & 0.7572 & 0.7932 & 0.7496 & - & - & {\color[HTML]{5B9BD5} {\ul 0.7956}} & {\color[HTML]{FF0000} \textbf{0.8001}} & 0.7835 \\
\multirow{-3}{*}{UHDM} & LPIPS$\downarrow$ & 0.5314 & 0.3764 & 0.3409 & 0.4882 & 0.3857 & 0.3318 & 0.3519 & - & - & 0.2551 & {\color[HTML]{5B9BD5} {\ul 0.2536}} & {\color[HTML]{FF0000} \textbf{0.2385}} \\ \hline
 & PSNR$\uparrow$ & 17.974 & 21.538 & 20.831 & - & 22.756 & 22.309 & 22.930 & 23.484 & 23.627 & 24.500 & {\color[HTML]{5B9BD5} {\ul 24.823}} & {\color[HTML]{FF0000} \textbf{24.941}} \\
 & SSIM$\uparrow$ & 0.7033 & 0.7727 & 0.7343 & - & 0.7958 & 0.8095 & 0.7885 & 0.8263 & 0.804 & 0.8351 & {\color[HTML]{FF0000} \textbf{0.8426}} & {\color[HTML]{5B9BD5} {\ul 0.8391}} \\
\multirow{-3}{*}{FHDMi} & LPIPS$\downarrow$ & 0.2837 & 0.2477 & 0.2515 & - & 0.1794 & 0.1980 & 0.1688 & - & 0.161 & {\color[HTML]{5B9BD5} {\ul 0.1354}} & {\color[HTML]{FF0000} \textbf{0.1288}} & 0.1396 \\ \hline
 & PSNR$\uparrow$ & 20.30 & 26.77 & - & 28.08 & 27.75 & 30.03 & 27.78 & {\color[HTML]{5B9BD5} {\ul 30.84}} & {\color[HTML]{FF0000} \textbf{30.94}} & 29.81 & 30.13* & 29.60 \\
\multirow{-2}{*}{TIP2018} & SSIM$\uparrow$ & 0.738 & 0.871 & - & 0.904 & 0.895 & 0.893 & 0.896 & 0.901 & 0.914 & {\color[HTML]{5B9BD5} {\ul 0.916}} & {\color[HTML]{FF0000} \textbf{0.920*}} & 0.914 \\ \hline
 & Params(M)$\downarrow$ & - & {\color[HTML]{5B9BD5} {\ul 1.426}} & 7.637 & 3.360 & 58.565 & 14.192 & 13.571 & {\color[HTML]{FF0000} \textbf{0.311}} & 15.4 & 5.934 & 5.662*/6.181 & 1.979 \\
 & FLOPs(G)$\downarrow$ & - & 561.48 & 917.29 & 436.85 & 6306.89 & 2110.89 & 8463.59 & {\color[HTML]{FF0000} \textbf{153.00}} & - & 559.96 & 1725.61 & {\color[HTML]{5B9BD5} {\ul 191.35}} \\
\multirow{-3}{*}{-} & Runtime(ms)$\downarrow$ & - & 80.55 & 152.20 & {\color[HTML]{5B9BD5} {\ul 75.02}} & 1646.68 & 270.62 & 1657.76 & 502.98 & - & 129.38 & 420.14 & {\color[HTML]{FF0000} \textbf{68.22}} \\ \hline
\end{tabular}}
\end{table*}

MFMAF fuses a concatenation of multiple features by weighting and then compressing them subtly and efficiently. 3D features can be divided into the channel dimension and spatial dimensions. Weighting the features can be done by calculating a 3D weight vector or by calculating weights separately on the channel and spatial dimensions. Considering the significant computational complexity of the former approach, we choose the latter one. When computing on channel dimension, the features are first globally average-pooled to a tensor with both width and height equal to 1. Then, an IdentityFormer Block (IFB) \cite{meta} is used to calculate the weights, which has been shown to have a solid lower bound of performance in computer vision tasks and compared to MLP, which is utilized by SENet, it adds almost no additional parameters or computational complexity.
When computing on spatial dimensions, the features go through a convolutional branch to obtain a more refined spatial representation with a reduced number of channels. Additionally, an average pooling branch is used to perform channel-wise pooling on each feature, resulting in the channels of pooled features being the same with the number of input features. Then outputs of two branches are concatenated and passed through a convolution layer to generate spatial attention weights, the channels of which are the same with the number of input features. Each feature has its own dedicated spatial attention weights. In the process of computing spatial attention, feature resolution is not interpolated to a fixed value, because this operation makes the scales of features chaotic, as a result, MFMAF is resolution-friendly.

From the above operations, MFMAF allows the network to capture the most relevant and informative aspects of each feature implicitly and adaptively.

\subsection{Loss Functions}
Just like most previous works, we employ pixel loss to train our model. However, due to lens distortion of cameras, moiré pattern images may have slight deformations and misalignments compared to clear images, making pixel loss insufficient. Therefore, we need to incorporate perceptual loss \cite{per} for feature-based supervision. The output images from all decoder levels are supervised during training. Hence, the final loss function is formulated as follows,
\begin{equation}
L = \sum_{i=1}^{3}\lambda_1\times L_1(GT_i,O_i) + \lambda_p\times L_p(GT_i,O_i),
\label{eq:l}
\end{equation}
where $L_1$ and $L_p$ represent the L1 pixel loss and perceptual loss, respectively. $\lambda_1$ and $\lambda_p$ are weighting factors to balance the importance of two losses and both are set to 1, which follows \cite{esd}. $O_i$ and $GT_i$ are output image of $DB_i$ and ground truth image with corresponding size, respectively.
For the perceptual loss, we utilize a pre-trained VGG16 network \cite{vgg} to extract features and compute their L1 distance in the feature space, which can be formulated as
\begin{equation}
L_p(GT_i,O_i) = \sum_{i=1}^{n}L_1(VGG16(GT_i,n),VGG16(O_i,n)),
\label{eq:lp}
\end{equation}
where $n$ indicates that the nth feature in a feature set selected from the VGG16 network is generated. By combining both pixel loss and perceptual loss, we can effectively guide the model to generate visually pleasing moiré-free images with high fidelity as well.

\section{EXPERIMENTAL RESULTS}
\label{sec:exp}
\begin{figure*}[htb]

\begin{minipage}[b]{0.12\linewidth}
\centerline{\includegraphics[width=2.125cm]{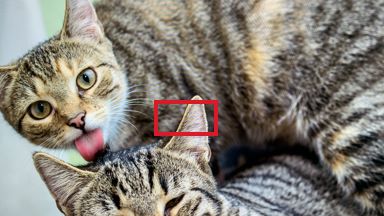}}
\centerline{\includegraphics[width=2.125cm]{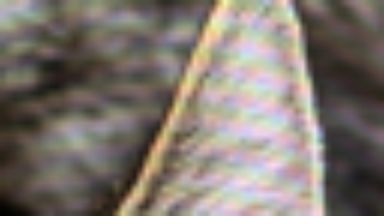}}
\vspace{0.1cm}
\centerline{\includegraphics[width=2.125cm]{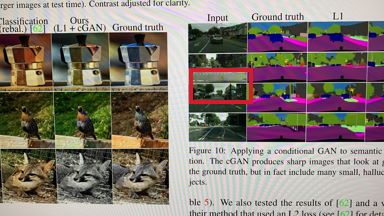}}
\centerline{\includegraphics[width=2.125cm]{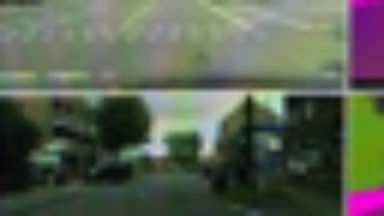}}
\vspace{0.1cm}
\centerline{\includegraphics[width=2.125cm]{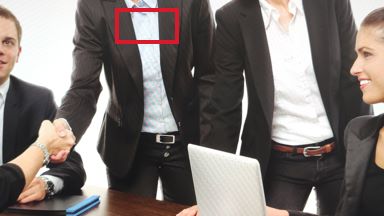}}
\centerline{\includegraphics[width=2.125cm]{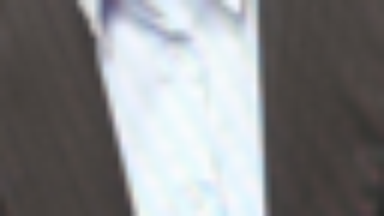}}
\centerline{(a)Input}\medskip
\end{minipage}
\begin{minipage}[b]{0.12\linewidth}
\centerline{\includegraphics[width=2.125cm]{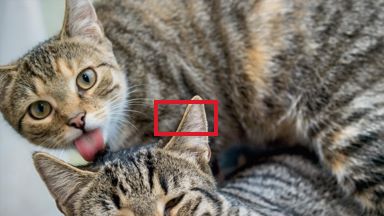}}
\centerline{\includegraphics[width=2.125cm]{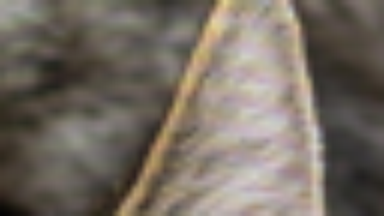}}
\vspace{0.1cm}
\centerline{\includegraphics[width=2.125cm]{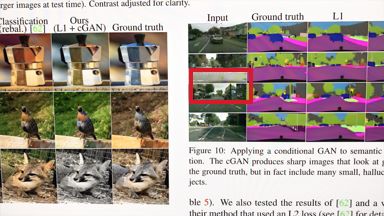}}
\centerline{\includegraphics[width=2.125cm]{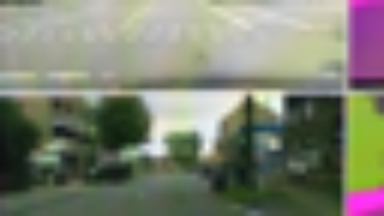}}
\vspace{0.1cm}
\centerline{\includegraphics[width=2.125cm]{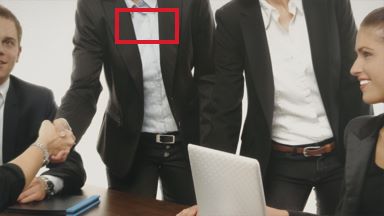}}
\centerline{\includegraphics[width=2.125cm]{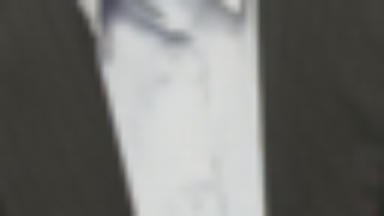}}
\centerline{(b)DMCNN}\medskip
\end{minipage}
\begin{minipage}[b]{0.12\linewidth}
\centerline{\includegraphics[width=2.125cm]{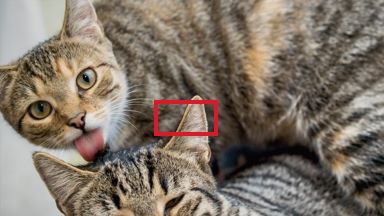}}
\centerline{\includegraphics[width=2.125cm]{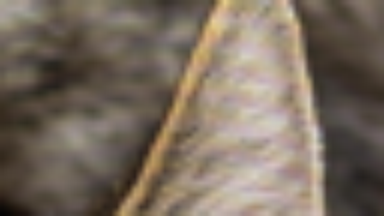}}
\vspace{0.1cm}
\centerline{\includegraphics[width=2.125cm]{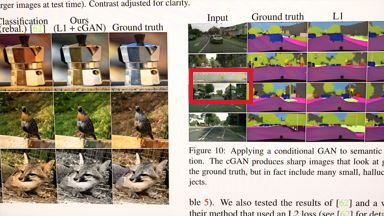}}
\centerline{\includegraphics[width=2.125cm]{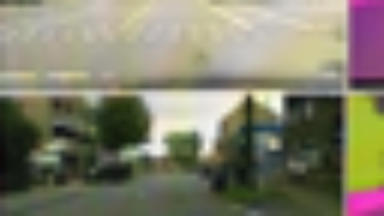}}
\vspace{0.1cm}
\centerline{\includegraphics[width=2.125cm]{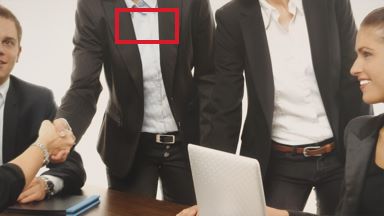}}
\centerline{\includegraphics[width=2.125cm]{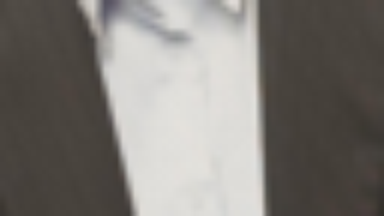}}
\centerline{(c)WDNet}\medskip
\end{minipage}
\begin{minipage}[b]{0.12\linewidth}
\centerline{\includegraphics[width=2.125cm]{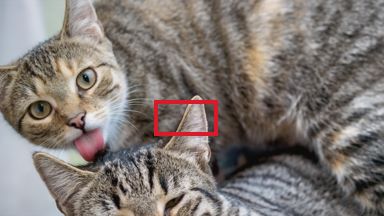}}
\centerline{\includegraphics[width=2.125cm]{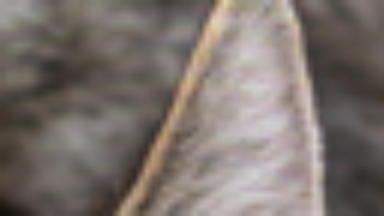}}
\vspace{0.1cm}
\centerline{\includegraphics[width=2.125cm]{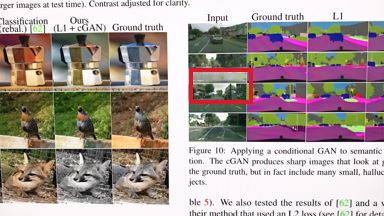}}
\centerline{\includegraphics[width=2.125cm]{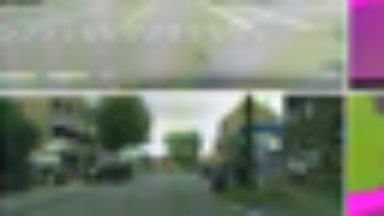}}
\vspace{0.1cm}
\centerline{\includegraphics[width=2.125cm]{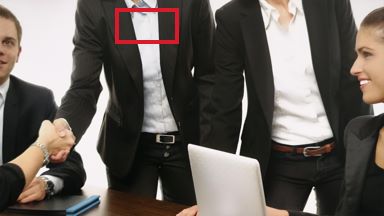}}
\centerline{\includegraphics[width=2.125cm]{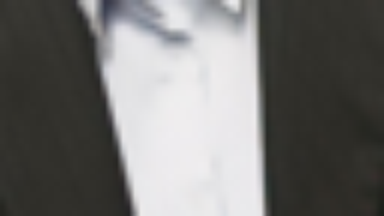}}
\centerline{(d)MBCNN}\medskip
\end{minipage}
\begin{minipage}[b]{0.12\linewidth}
\centerline{\includegraphics[width=2.125cm]{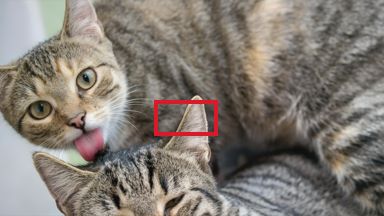}}
\centerline{\includegraphics[width=2.125cm]{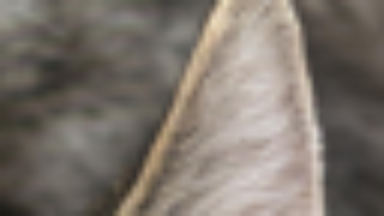}}
\vspace{0.1cm}
\centerline{\includegraphics[width=2.125cm]{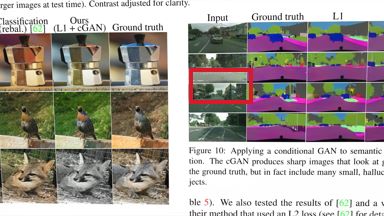}}
\centerline{\includegraphics[width=2.125cm]{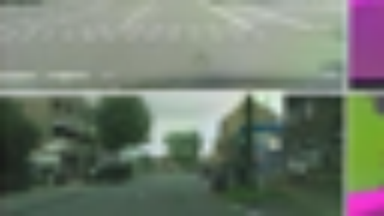}}
\vspace{0.1cm}
\centerline{\includegraphics[width=2.125cm]{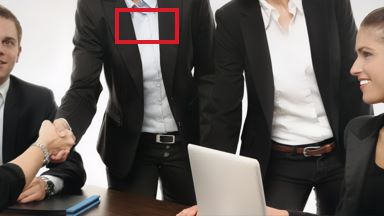}}
\centerline{\includegraphics[width=2.125cm]{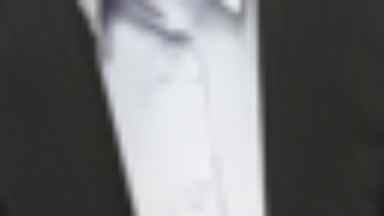}}
\centerline{(e)ESDNet}\medskip
\end{minipage}
\begin{minipage}[b]{0.12\linewidth}
\centerline{\includegraphics[width=2.125cm]{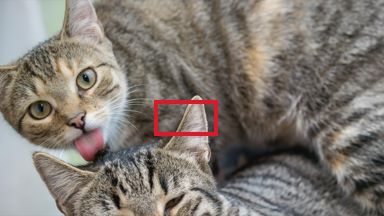}}
\centerline{\includegraphics[width=2.125cm]{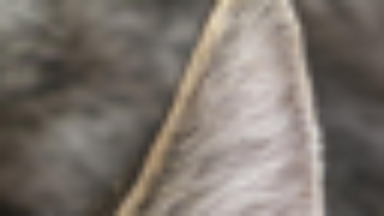}}
\vspace{0.1cm}
\centerline{\includegraphics[width=2.125cm]{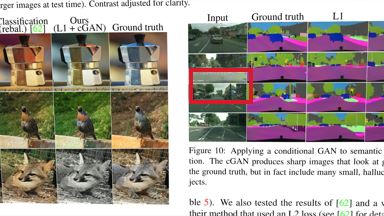}}
\centerline{\includegraphics[width=2.125cm]{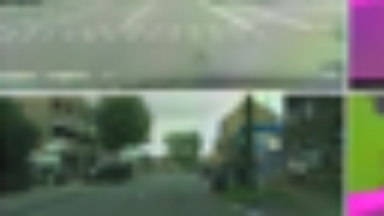}}
\vspace{0.1cm}
\centerline{\includegraphics[width=2.125cm]{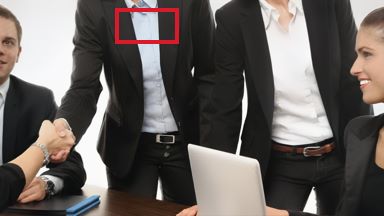}}
\centerline{\includegraphics[width=2.125cm]{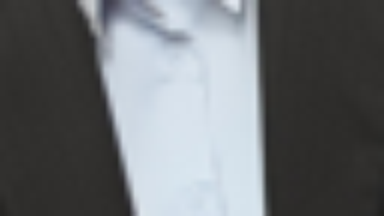}}
\centerline{(f)MCFNet}\medskip
\end{minipage}
\begin{minipage}[b]{0.12\linewidth}
\centerline{\includegraphics[width=2.125cm]{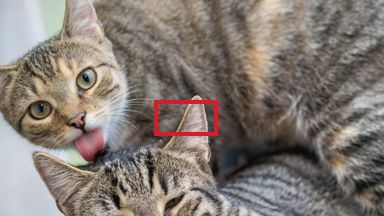}}
\centerline{\includegraphics[width=2.125cm]{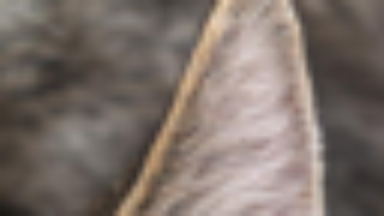}}
\vspace{0.1cm}
\centerline{\includegraphics[width=2.125cm]{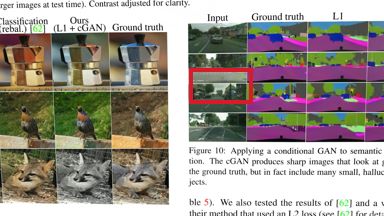}}
\centerline{\includegraphics[width=2.125cm]{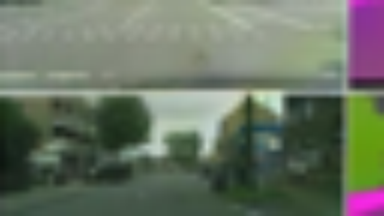}}
\vspace{0.1cm}
\centerline{\includegraphics[width=2.125cm]{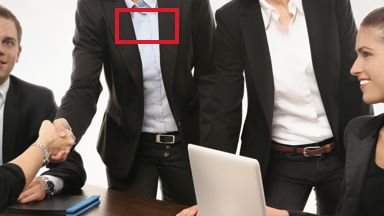}}
\centerline{\includegraphics[width=2.125cm]{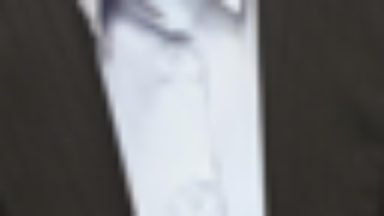}}
\centerline{(g)Ours}\medskip
\end{minipage}
\begin{minipage}[b]{0.12\linewidth}
\centerline{\includegraphics[width=2.125cm]{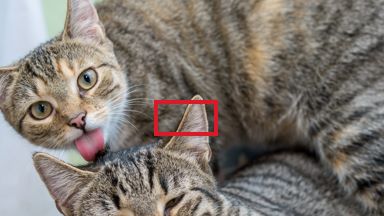}}
\centerline{\includegraphics[width=2.125cm]{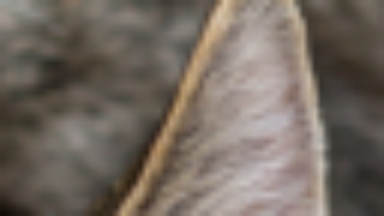}}
\vspace{0.1cm}
\centerline{\includegraphics[width=2.125cm]{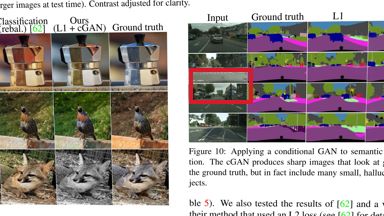}}
\centerline{\includegraphics[width=2.125cm]{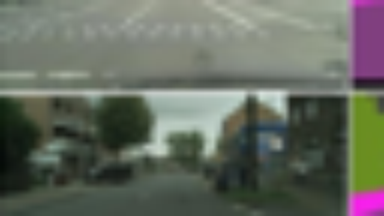}}
\vspace{0.1cm}
\centerline{\includegraphics[width=2.125cm]{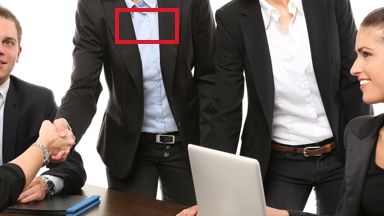}}
\centerline{\includegraphics[width=2.125cm]{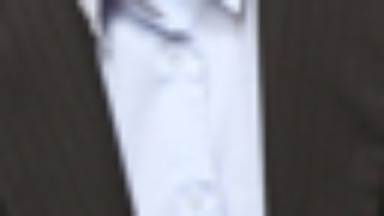}}
\centerline{(h)GT}\medskip
\end{minipage}

\caption{Qualitative comparison with classic methods on the UHDM dataset. Red boxes show zoom-in regions for demonstrating better details.}
\label{fig:exp}
\end{figure*} 
\subsection{Datasets and Implementation Details}
We evaluate the proposed method on three real-world datasets specifically designed for screenshot demoiréing: TIP2018 \cite{dm}, FHDMi \cite{fhd}, and UHDM \cite{esd}, and compare it with state-of-the-art methods quantitatively and qualitatively. These datasets consist of photographs with authentic moiré patterns captured by cameras, showcasing a wide range of contents and diverse moiré pattern styles. The FHDMi dataset contains images with a FHD ($1920\times1080$) resolution, while the UHDM dataset contains images with a UHD ($3840\times2160$) resolution, which closely reflects real-world application scenarios. Yu et al. \cite{esd} use multiple cameras to capture multiple screens from different positions to generate the dataset, which makes UHDM cover varying real-world conditions.

We implement the proposed framework with PyTorch, on a single NVIDIA RTX 3090 GPU. Adam is used as our training optimizer with $\beta_1=0.9$ and $\beta_2=0.999$. 
The learning rate is initialized to be $2\times10^{-4}$, 
and we employ CosineAnnealingWarmRestarts \cite{cos} as the learning rate scheduler with $T_0=20$, $T_{mult}=1$ and $eta_{min}=1\times10^{-6}$. 
We divide the training process into two stages: In the first stage, we crop the image pairs to a middle size to speed up training and train the model for $3\times T_0$ cycles. 
In the second stage, we crop the image pairs to a larger size and train the model until the sequence of training losses in every last epoch of a $T_0$ cycle is no longer decreasing to adjust the training epochs adaptively.

\subsection{Quantitative Analysis}
Table \ref{table:1} presents a comparison of existing demoiréing methods based on three quantitative performance metrics: PSNR, SSIM, and LPIPS \cite{lpips}. PMTNet and CFDNet have not been trained on the UHDM dataset, and because of their ordinary performance on the FHDMi dataset, which has a similar characteristic and resolution to the UHDM dataset, it is reasonable to have low expectations for their performance on the UHDM dataset. 

The proposed method demonstrates state-of-the-art results on certain metrics and datasets while achieving comparable results on others. Meanwhile, FC3DNet has the lowest runtime and the second lowest FLOPs, which are only one-seventh and one-ninth compared to the runtime and FLOPs of MCFNet, respectively. Limited capabilities of our model caused by its relatively fewer parameters, prevent it from excelling in all aspects. However, It matches the intention of proposing this model. Notably, the proposed method exhibits superior performance on the UHDM dataset and FHDMi dataset compared to the TIP2018 dataset, images of which have lower resolutions. This highlights the effectiveness of our method in handling high-resolution moiré pattern images, making it suitable for real-world demoiréing scenarios.
\subsection{Qualitative Analysis}
\begin{figure}[t]
\centering
\begin{minipage}[b]{0.48\linewidth}
 \centering
 \centerline{\includegraphics[width=8.5 cm]{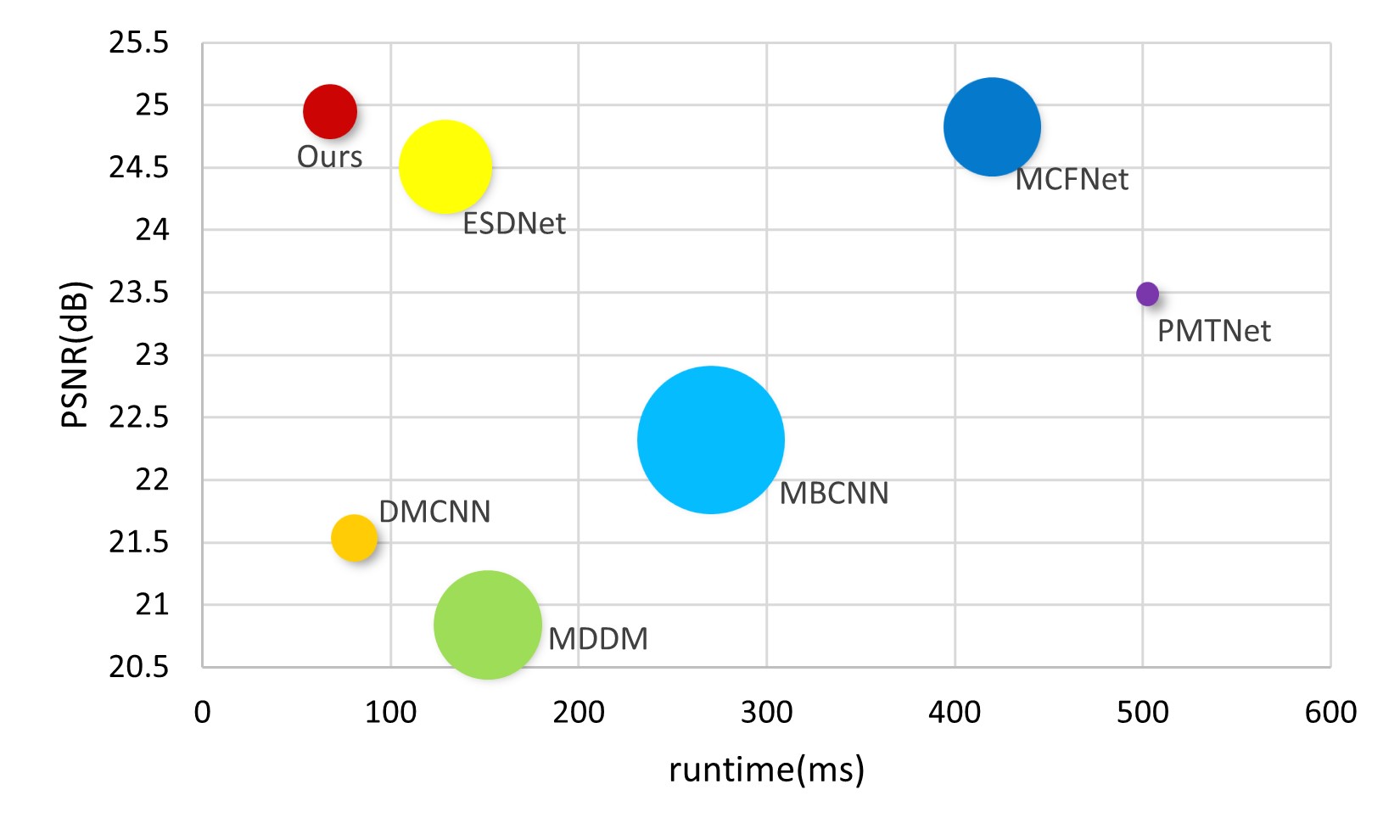}}
\end{minipage}
\caption{Comparison on efficiency. PSNR and runtime are tested on the FHDMi dataset using a RTX 3080 GPU. Areas of circles denote the parameter number of networks. MopNet and FHD$^2$eNet are exclusive due to ordinary performance and overlong runtime.}
\label{fig:ef}
\end{figure}
Fig. \ref{fig:exp} qualitatively compares the results obtained by our method with other methods on the UHDM dataset. In Fig. \ref{fig:exp}, DMCNN, WDNet, and MBCNN, which were developed before the introduction of FHDMi, struggle to remove large moiré patterns in the first and second images and have a loss of details in the third image. The proposed method, along with ESDNet and MCFNet, produces better demoiréing results. However, these methods retain slight green moiré stripes in the second image and MCFNet has a little loss of details on the second button of the shirt. Although the proposed method successfully removes the majority of moiré patterns while preserving a large amount of fine details, there are still some highly complex moiré patterns that are difficult to remove completely.
\subsection{Efficiency Study}
In addition to the quantitative and qualitative results, our method outperforms all existing approaches in terms of efficiency. As shown in Fig. \ref{fig:ef}, our method achieves a state-of-the-art result on the FHDMi dataset while utilizing the least runtime. Indeed, it is not always the case that reducing the number of network parameters leads to faster network speeds. The network’s architecture also plays a crucial role in determining its efficiency. 

\subsection{Ablation Study}
\begin{table}[t]
\setlength{\tabcolsep}{3pt}
\centering
\caption{Ablation results on the overall architecture and fusion module. We design different networks to have similar parameters for fairness.}
\label{table:2}
\scalebox{0.9}{
\begin{tabular}{c|c|ccc}
\hline
Overall Architecture & Fusion Module & PSNR$\uparrow$ & SSIM$\uparrow$ & LPIPS$\downarrow$ \\ \hline
UNet-like & CBAM & 21.891 & 0.7636 & 0.2968 \\ \hline
FC3 & CBAM & 22.062 & 0.7729 & {\color[HTML]{5B9BD5} {\ul 0.2594}} \\ \hline
UNet-like & MFMAF & {\color[HTML]{5B9BD5} {\ul 22.148}} & {\color[HTML]{FF0000} \textbf{0.7873}} & 0.2726 \\ \hline
FC3 & MFMAF & {\color[HTML]{FF0000} \textbf{22.487}} & {\color[HTML]{5B9BD5} {\ul 0.7835}} & {\color[HTML]{FF0000} \textbf{0.2385}} \\ \hline
\end{tabular}}
\end{table}
We conduct an analysis on the UHDM dataset to evaluate the effectiveness of our proposed fully connected encoder-decoder (FC3) architecture, as well as the MFMAF module. 

In comparison to the UNet-like architecture, which is used as the overall architecture by some demoiréing networks \cite{mb,esd}, FC3 outperforms it in terms of PSNR and LPIPS, which measures the perceptual similarity of two images, indicating better visually pleasing images. Indeed, allowing each level of the decoder to utilize the outputs from all levels of the encoder enables our network to gather more useful information and ultimately achieve better visual perception. 

When compared to CBAM \cite{cbam}, the Convolutional Block Attention Module, which is widely used, MFMAF significantly enhances the network’s ability to remove moiré patterns. This indicates that MFMAF is stronger than CBAM in terms of fusing multiple features and is better suited for our architecture's new requirement for feature fusion.

\section{CONCLUSIONS}
\label{sec:conclu}
We propose a novel fully connected encoder-decoder based demoiréing network, which makes full use of multi-scale features to effectively and efficiently remove global and local moiré patterns in screenshot photos. Specifically, features from all levels of the encoder are connected and fed into each level of the decoder. Multi-Feature Multi-Attention Fusion module is designed to fuse features with multiple depths and receptive fields into one feature for further processing. 
Experimental results on three real-world datasets demonstrate our method achieves comparable or even superior performance compared to state-of-the-art methods while utilizing only a fraction of parameters, FLOPs, and runtime. Thus, we have achieved a better balance between the network speed and the quality of results.
An important direction for future research is to optimize and deploy our method on mobile devices to further validate its practicality in real-world scenarios.

\vfill\pagebreak

\bibliographystyle{IEEEbib}
\bibliography{refs}

\begin{thebibliography}{10}

\bibitem{wang2023decoupling}
Yang Wang, Long Peng, Liang Li, Yang Cao, and Zheng-Jun Zha,
\newblock ``Decoupling-and-aggregating for image exposure correction,''
\newblock in {\em Proceedings of the IEEE/CVF Conference on Computer Vision and Pattern Recognition}, 2023, pp. 18115--18124.

\bibitem{wang2023brightness}
Haodian Wang, Long Peng, Yuejin Sun, Zengyu Wan, Yang Wang, and Yang Cao,
\newblock ``Brightness perceiving for recursive low-light image enhancement,''
\newblock {\em IEEE Transactions on Artificial Intelligence}, 2023.

\bibitem{peng2024lightweight}
Long Peng, Yang Cao, Yuejin Sun, and Yang Wang,
\newblock ``Lightweight adaptive feature de-drifting for compressed image classification,''
\newblock {\em IEEE Transactions on Multimedia}, 2024.

\bibitem{peng2024towards}
Long Peng, Wenbo Li, Renjing Pei, Jingjing Ren, Yang Wang, Yang Cao, and Zheng-Jun Zha,
\newblock ``Towards realistic data generation for real-world super-resolution,''
\newblock {\em arXiv preprint arXiv:2406.07255}, 2024.

\bibitem{peng2024efficient}
Long Peng, Yang Cao, Renjing Pei, Wenbo Li, Jiaming Guo, Xueyang Fu, Yang Wang, and Zheng-Jun Zha,
\newblock ``Efficient real-world image super-resolution via adaptive directional gradient convolution,''
\newblock {\em arXiv preprint arXiv:2405.07023}, 2024.

\bibitem{wang2017deep}
Yang Wang, Jing Zhang, Yang Cao, and Zengfu Wang,
\newblock ``A deep cnn method for underwater image enhancement,''
\newblock in {\em 2017 IEEE international conference on image processing (ICIP)}. IEEE, 2017, pp. 1382--1386.

\bibitem{li2023ntire}
Yawei Li, Yulun Zhang, Radu Timofte, Luc Van~Gool, Lei Yu, Youwei Li, Xinpeng Li, Ting Jiang, Qi~Wu, Mingyan Han, et~al.,
\newblock ``Ntire 2023 challenge on efficient super-resolution: Methods and results,''
\newblock in {\em Proceedings of the IEEE/CVF Conference on Computer Vision and Pattern Recognition}, 2023, pp. 1921--1959.

\bibitem{wang2020deep}
Yang Wang, Yang Cao, Zheng-Jun Zha, Jing Zhang, and Zhiwei Xiong,
\newblock ``Deep degradation prior for low-quality image classification,''
\newblock in {\em Proceedings of the IEEE/CVF Conference on Computer Vision and Pattern Recognition}, 2020, pp. 11049--11058.

\bibitem{wang2019progressive}
Yang Wang, Yang Cao, Zheng-Jun Zha, Jing Zhang, Zhiwei Xiong, Wei Zhang, and Feng Wu,
\newblock ``Progressive retinex: Mutually reinforced illumination-noise perception network for low-light image enhancement,''
\newblock in {\em Proceedings of the 27th ACM international conference on multimedia}, 2019, pp. 2015--2023.

\bibitem{peng2021ensemble}
Long Peng, Aiwen Jiang, Haoran Wei, Bo~Liu, and Mingwen Wang,
\newblock ``Ensemble single image deraining network via progressive structural boosting constraints,''
\newblock {\em Signal Processing: Image Communication}, vol. 99, pp. 116460, 2021.

\bibitem{peng2020cumulative}
Long Peng, Aiwen Jiang, Qiaosi Yi, and Mingwen Wang,
\newblock ``Cumulative rain density sensing network for single image derain,''
\newblock {\em IEEE Signal Processing Letters}, vol. 27, pp. 406--410, 2020.

\bibitem{zou2022dreaming}
Weiqi Zou, Yang Wang, Xueyang Fu, and Yang Cao,
\newblock ``Dreaming to prune image deraining networks,''
\newblock in {\em Proceedings of the IEEE/CVF Conference on Computer Vision and Pattern Recognition}, 2022, pp. 6023--6032.

\bibitem{mop}
Bin He, Ce~Wang, Boxin Shi, and Ling-Yu Duan,
\newblock ``Mop moire patterns using mopnet,''
\newblock in {\em Proceedings of the IEEE/CVF International Conference on Computer Vision}, 2019, pp. 2424--2432.

\bibitem{cfd}
Ce~Wang, Bin He, Shengsen Wu, Renjie Wan, Boxin Shi, and Ling-Yu Duan,
\newblock ``Coarse-to-fine disentangling demoir{\'e}ing framework for recaptured screen images,''
\newblock {\em IEEE Transactions on Pattern Analysis and Machine Intelligence}, 2023.

\bibitem{fhd}
Bin He, Ce~Wang, Boxin Shi, and Ling-Yu Duan,
\newblock ``Fhde 2 net: Full high definition demoireing network,''
\newblock in {\em Computer Vision--ECCV 2020: 16th European Conference, Glasgow, UK, August 23--28, 2020, Proceedings, Part XXII 16}. Springer, 2020, pp. 713--729.

\bibitem{dm}
Yujing Sun, Yizhou Yu, and Wenping Wang,
\newblock ``Moir{\'e} photo restoration using multiresolution convolutional neural networks,''
\newblock {\em IEEE Transactions on Image Processing}, vol. 27, no. 8, pp. 4160--4172, 2018.

\bibitem{md}
Xi~Cheng, Zhenyong Fu, and Jian Yang,
\newblock ``Multi-scale dynamic feature encoding network for image demoir{\'e}ing,''
\newblock in {\em 2019 IEEE/CVF International Conference on Computer Vision Workshop (ICCVW)}. IEEE, 2019, pp. 3486--3493.

\bibitem{pmt}
Yuzhen Niu, Zhihua Lin, Wenxi Liu, and Wenzhong Guo,
\newblock ``Progressive moire removal and texture complementation for image demoireing,''
\newblock {\em IEEE Transactions on Circuits and Systems for Video Technology}, 2023.

\bibitem{sdn}
Duong~Hai Nguyen and Chul Lee,
\newblock ``A contrastive learning approach for screenshot demoir{\'e}ing,''
\newblock in {\em 2023 IEEE International Conference on Image Processing (ICIP)}. IEEE, 2023, pp. 1210--1214.

\bibitem{mcf}
Duong~Hai Nguyen, Se-Ho Lee, and Chul Lee,
\newblock ``Multiscale coarse-to-fine guided screenshot demoir{\'e}ing,''
\newblock {\em IEEE Signal Processing Letters}, 2023.

\bibitem{u}
Olaf Ronneberger, Philipp Fischer, and Thomas Brox,
\newblock ``U-net: Convolutional networks for biomedical image segmentation,''
\newblock in {\em Medical Image Computing and Computer-Assisted Intervention--MICCAI 2015: 18th International Conference, Munich, Germany, October 5-9, 2015, Proceedings, Part III 18}. Springer, 2015, pp. 234--241.

\bibitem{mb}
Bolun Zheng, Shanxin Yuan, Gregory Slabaugh, and Ales Leonardis,
\newblock ``Image demoireing with learnable bandpass filters,''
\newblock in {\em Proceedings of the IEEE/CVF Conference on Computer Vision and Pattern Recognition}, 2020, pp. 3636--3645.

\bibitem{esd}
Xin Yu, Peng Dai, Wenbo Li, Lan Ma, Jiajun Shen, Jia Li, and Xiaojuan Qi,
\newblock ``Towards efficient and scale-robust ultra-high-definition image demoir{\'e}ing,''
\newblock in {\em European Conference on Computer Vision}. Springer, 2022, pp. 646--662.

\bibitem{se}
Jie Hu, Li~Shen, and Gang Sun,
\newblock ``Squeeze-and-excitation networks,''
\newblock in {\em Proceedings of the IEEE conference on computer vision and pattern recognition}, 2018, pp. 7132--7141.

\bibitem{wd}
Lin Liu, Jianzhuang Liu, Shanxin Yuan, Gregory Slabaugh, Ale{\v{s}} Leonardis, Wengang Zhou, and Qi~Tian,
\newblock ``Wavelet-based dual-branch network for image demoir{\'e}ing,''
\newblock in {\em Computer Vision--ECCV 2020: 16th European Conference, Glasgow, UK, August 23--28, 2020, Proceedings, Part XIII 16}. Springer, 2020, pp. 86--102.

\bibitem{cbam}
Sanghyun Woo, Jongchan Park, Joon-Young Lee, and In~So Kweon,
\newblock ``Cbam: Convolutional block attention module,''
\newblock in {\em Proceedings of the European conference on computer vision (ECCV)}, 2018, pp. 3--19.

\bibitem{ps}
Wenzhe Shi, Jose Caballero, Ferenc Husz{\'a}r, Johannes Totz, Andrew~P Aitken, Rob Bishop, Daniel Rueckert, and Zehan Wang,
\newblock ``Real-time single image and video super-resolution using an efficient sub-pixel convolutional neural network,''
\newblock in {\em Proceedings of the IEEE conference on computer vision and pattern recognition}, 2016, pp. 1874--1883.

\bibitem{meta}
Weihao Yu, Chenyang Si, Pan Zhou, Mi~Luo, Yichen Zhou, Jiashi Feng, Shuicheng Yan, and Xinchao Wang,
\newblock ``Metaformer baselines for vision,''
\newblock {\em IEEE Transactions on Pattern Analysis and Machine Intelligence}, 2023.

\bibitem{per}
Justin Johnson, Alexandre Alahi, and Li~Fei-Fei,
\newblock ``Perceptual losses for real-time style transfer and super-resolution,''
\newblock in {\em Computer Vision--ECCV 2016: 14th European Conference, Amsterdam, The Netherlands, October 11-14, 2016, Proceedings, Part II 14}. Springer, 2016, pp. 694--711.

\bibitem{vgg}
Karen Simonyan and Andrew Zisserman,
\newblock ``Very deep convolutional networks for large-scale image recognition,''
\newblock {\em arXiv preprint arXiv:1409.1556}, 2014.

\bibitem{cos}
Ilya Loshchilov and Frank Hutter,
\newblock ``Sgdr: Stochastic gradient descent with warm restarts,''
\newblock {\em arXiv preprint arXiv:1608.03983}, 2016.

\bibitem{lpips}
Richard Zhang, Phillip Isola, Alexei~A Efros, Eli Shechtman, and Oliver Wang,
\newblock ``The unreasonable effectiveness of deep features as a perceptual metric,''
\newblock in {\em Proceedings of the IEEE conference on computer vision and pattern recognition}, 2018, pp. 586--595.

\end{thebibliography}

\end{document}